\documentclass[10pt,twocolumn,letterpaper]{article}

\usepackage{cvpr}              
\usepackage[accsupp]{axessibility} 

\usepackage[dvipsnames]{xcolor}

\definecolor{cvprblue}{rgb}{0.21,0.49,0.74}
\usepackage[pagebackref,breaklinks,colorlinks,citecolor=cvprblue]{hyperref}
\setlist[itemize]{noitemsep, topsep=0pt}

\usepackage{hyperref}
\usepackage{url}
\usepackage{graphicx}
\usepackage{booktabs}
\usepackage{enumitem}
\usepackage{multirow}
\usepackage{wrapfig}
\usepackage{subcaption}
\usepackage{amssymb}
\usepackage{pifont}
\newcommand{\cmark}{\ding{51}}%
\newcommand{\xmark}{\ding{55}}%

\def\rve{{\mathbf{e}}}

\def\rvs{{\mathbf{s}}}

\def\rvw{{\mathbf{w}}}
\def\rvx{{\mathbf{x}}}
\def\rvy{{\mathbf{y}}}
\def\rvz{{\mathbf{z}}}

\def\paperID{14386} 
\def\confName{CVPR}
\def\confYear{2024}
\begin{document}
\title{Customization Assistant for Text-to-image Generation}

\author{Yufan Zhou,~~ Ruiyi Zhang,~~ Jiuxiang Gu,~~ Tong Sun\\
Adobe Research \\
{\tt\small \{yufzhou, ruizhang, jigu, tsun\}@adobe.com}
}

\maketitle
\begin{abstract}
Customizing pre-trained text-to-image generation model has attracted massive research interest recently, due to its huge potential in real-world applications. Although existing methods are able to generate creative content for a novel concept contained in single user-input image, their capability are still far from perfection. Specifically, most existing methods require fine-tuning the generative model on testing images. Some existing methods do not require fine-tuning, while their performance are unsatisfactory. Furthermore, the interaction between users and models are still limited to directive and descriptive prompts such as instructions and captions. In this work, we build a customization assistant based on pre-trained large language model and diffusion model, which can not only perform customized generation in a tuning-free manner, but also enable more user-friendly interactions: users can chat with the assistant and input either ambiguous text or clear instruction. Specifically, we propose a new framework consists of a new model design and a novel training strategy. The resulting assistant can perform customized generation in 2-5 seconds without any test time fine-tuning. Extensive experiments are conducted, competitive results have been obtained across different domains, illustrating the effectiveness of the proposed method.
\end{abstract}    
\section{Introduction}
\label{sec:intro}
Customizing pre-trained text-to-image models has drawn significant interest in the research community, due to its  potential in real-world applications. Customized generation aims at generating creative images for a specific concept contained in user provided images. Despite of the impressive progress that large-scale text-to-image generation models have made in recent years~\citep{ramesh2021zero,ding2021cogview,rombach2022LDM,ramesh2022hierarchical, saharia2022imagen,yu2022scaling,chang2023muse,yu2023CM3Leon}, they fail to perform customized generation for novel concept, such as a specific animal or object which only appear in single testing image.

Various approaches have been proposed to tackle this task. Some methods focus on fine-tuning the pre-trained generation model~\citep{ruiz2023dreambooth,kumari2022multi,qiu2023oft} on testing images, so that the model can learn fine-grained details about the novel concept; Some methods aim to represent the concept by embeddings~\citep{gal2022textualinversion,wei2023elite,gal2023e4t,zhou2023profusion,li2023blipdiffusion}, which can be obtained either by optimization method or through a learned encoder. The embeddings are then injected into pre-trained text-to-image generation models to perform customized generation. The test time required for running these methods varies, spanning from few seconds to up to thirty minutes. We will discuss more details about related works in later section.

\begin{figure}[t!]
    \centering
    \includegraphics[width=0.95 \linewidth]{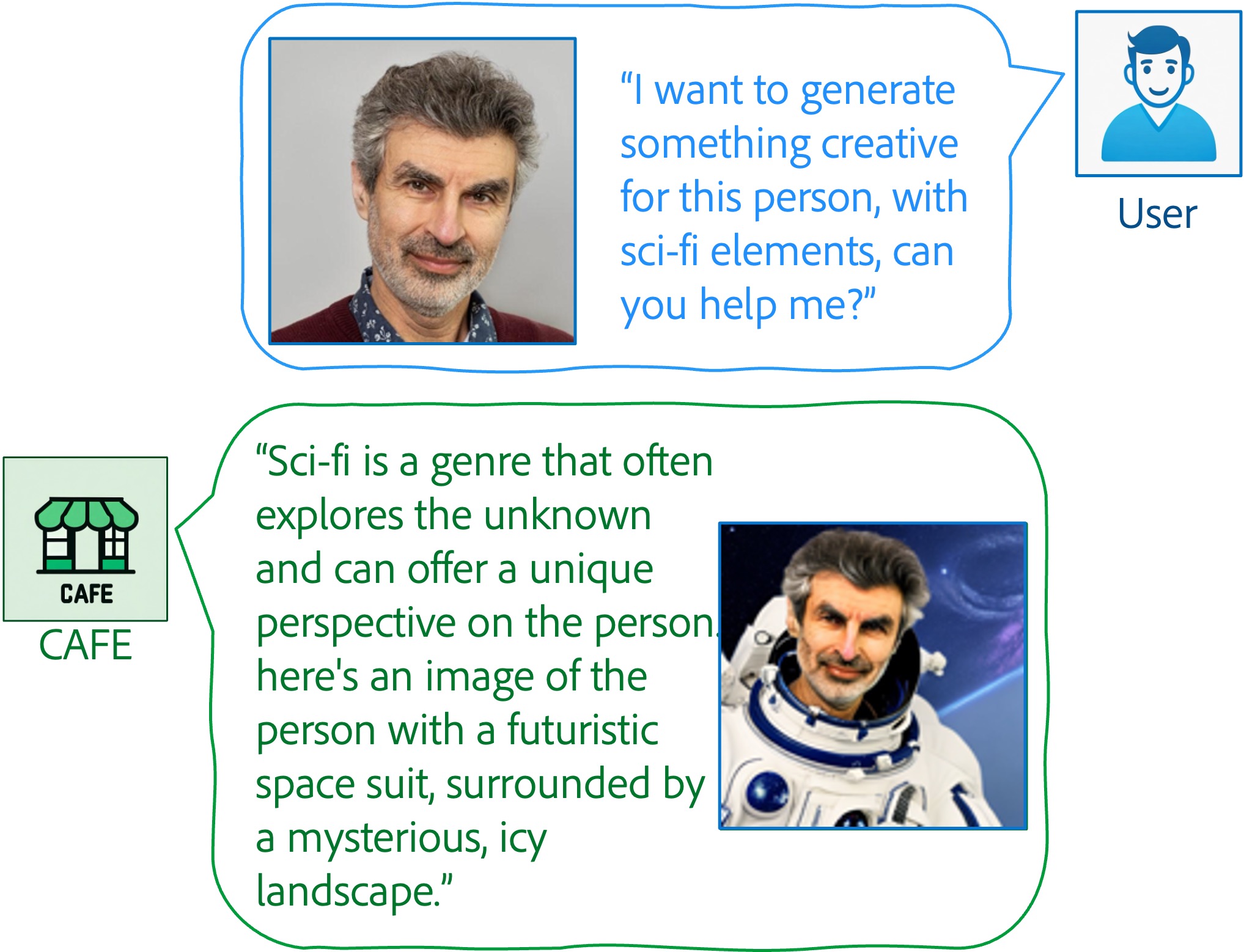}
    \caption{Generated example from the proposed CAFE. CAFE can perform customized generation based on the user provided image in a tuning-free manner. It outputs creative images along with text explanation and elaboration.}
    \label{fig:first_example}
    \vspace{-0.2in}
\end{figure}
Although existing methods are capable of generating creative contents for the target novel concept, they still have drawbacks and limitations.
For instance, all these existing methods are not user-friendly enough: they can only handle prompts that are directive or descriptive in nature, such as a caption ``A picture of the dog in comic book style" or an instruction ``Generate an image of the dog in comic book style". They can not handle ambiguous input such as ``I want to generate something creative, can you help me?", which could be important in applications because the user may only have a vague target instead of precise requirements in mind.

In this work, we propose CAFE which is a \textit{\textbf{C}ustomization \textbf{A}ssistant \textbf{F}or text-to-imag\textbf{E} generation}. CAFE is able to perform tuning-free customization within 2-5 seconds on testing images from arbitrary domain. Thus it is one of the most efficient customization method. 
Different from existing methods which take instructions or captions as text input, CAFE can handle both declarative and interrogative sentences because it is built upon a large language model (LLM) Llama-2~\citep{touvron2023llama2}. 
Furthermore, CAFE is user-friendly than any existing methods as it can even infer user's intention when the prompt is ambiguous, and output explanation and elaboration for the generation as shown in Figure~\ref{fig:first_example}.

Our contributions can be summarized as following:
\begin{itemize}
    \item We propose CAFE, a novel method which can perform tuning-free customized generation in 2-5 seconds. Different from previous methods, CAFE is built upon large language models, thus can handle ambiguous text input. Furthermore, CAFE can take extra images as additional semantic condition. It also possesses the unique capability of providing text explanation and elaboration for the generated content, which none of existing customization method can achieve;
    \item We propose a novel training strategy, which can efficiently construct large-scale high-quality dataset for training CAFE without human supervision thus saves huge amount of cost;
    \item Extensive experiments are conducted, where CAFE achieves promising quantitative and qualitative results across different domains. We also conduct several ablation studies, which verify the underlying rationale of the proposed method.
\end{itemize}

\section{Related Works}
\label{sec:related_works}
\paragraph{Text-to-image Generation}
The field of text-to-image generation has been a subject of research for years and has recently seen remarkable advancements. Previous methods which are based on Generative Adversarial Networks~\citep{xu2018attngan,tao2021dfgan,zhang2021cross,zhou2021lafite} lack the ability of generating open-domain images with arbitrary text input.
Starting from DALL-E~\citep{ramesh2021zero}, researchers are able to perform impressive zero-shot text-to-image generation with good fidelity and image-text alignment, after training the model on large-scale dataset. Specifically, DALL-E~\citep{ramesh2021zero} and CogView~\citep{ding2021cogview} propose to use transformer model to infer image tokens from text, which will be further transformed into images through an auto-encoder. 
GLIDE~\citep{nichol2021glide} adopts a hierarchical architecture which consists of diffusion models at different resolution, leading to impressive generation quality. The idea of using hierarchical design is also adopted by some follow-up works~\citep{ramesh2022hierarchical,chang2023muse} and proven to be effective in both diffusion models and auto-regressive models.
LDM~\citep{rombach2022LDM} proposes to train a diffusion model inside the lower-dimensional latent space of  auto-encoder, leading to better generation efficiency. 
Base on the research detailed above, numerous efforts have been undertaken to further enhance the proficiency of text-to-image generation models, including introducing better semantic understanding through a pre-trained large-scale text encoder~\citep{saharia2022imagen} and scaling up the model towards better generalization ability and more promising results~\citep{yu2022scaling,yu2023CM3Leon}.  In this work, we chose Stable Diffusion~\citep{rombach2022LDM} as the foundation for our CAFE due to its open-source availability.

\paragraph{Customized Generation} 
To enable customized generation for the specific concept presented in few-shot or single image, many customization methods proposed to fine-tune the pre-trained text-to-image generation model.
DreamBooth~\citep{ruiz2023dreambooth} propose to fine-tuned the entire diffusion model, while Custom Diffusion~\citep{kumari2023multi} only fine-tuned the cross attention module inside the UNet of diffusion model. LoRA~\citep{hu2021lora} is often used to reduce the number of parameters to be tuned and improve the efficiency of fine-tuning. Recently, OFT~\citep{qiu2023oft} is proposed, which can stabilize the fine-tuning process by preserving pairwise neuron relationship of pre-trained diffusion model.

Other works focus on representing target concept by learned embeddings.
Textual Inversion~\citep{gal2022textualinversion} propose to encode the concept by embedding vectors inside the input space of text encoder of the diffusion model. Optimization method is utilized to obtain the embedding, which may take up to 30 minutes. To reduce the time cost, different works~\citep{gal2023e4t,li2023blipdiffusion,zhou2023profusion,chen2023photoverse} have been proposed, focusing on pre-training encoders which can directly map the testing images into the target embeddings. Image patch features from pre-trained image encoders are often used to enhance the performance because some detailed information might be challenging to be captured inside the input embedding space of pre-trained text encoder~\citep{wei2023elite, shi2023instantbooth,chen2023anydoor,ma2023subjectdiffusion}.

There are also some other investigations: SuTI~\citep{chen2023suti} proposes to employ apprenticeship learning to obtain one single apprentice model to imitate half a million subject-specific experts; 
Kosmos-G~\citep{pan2023kosmos} tries to utilize large language model so that interleaved vision-language prompt can be handled; HyperDreamBooth~\citep{ruiz2023hyperdreambooth} directly obtains a customized model by training a hyper-network to generate the weights for target models. 
\section{Method}
\label{sec:method}
Our goal is to design an assistant which can generate creative images for a target object or person provided by single testing image in a tuning-free manner, the generated results should be aligned with arbitrary user-input text.
Different from existing works, we expect our assistant to be more user-friendly: the assistant should be able to handle both declarative and interrogative sentences; when the user-input is ambiguous, it should have the ability to infer user's intention and generate corresponding images; As an assistant, it is also expected to explain reason and insight behind its generated content through natural language;

\begin{figure}[t!]
    \centering
    \includegraphics[width=1.0\linewidth]{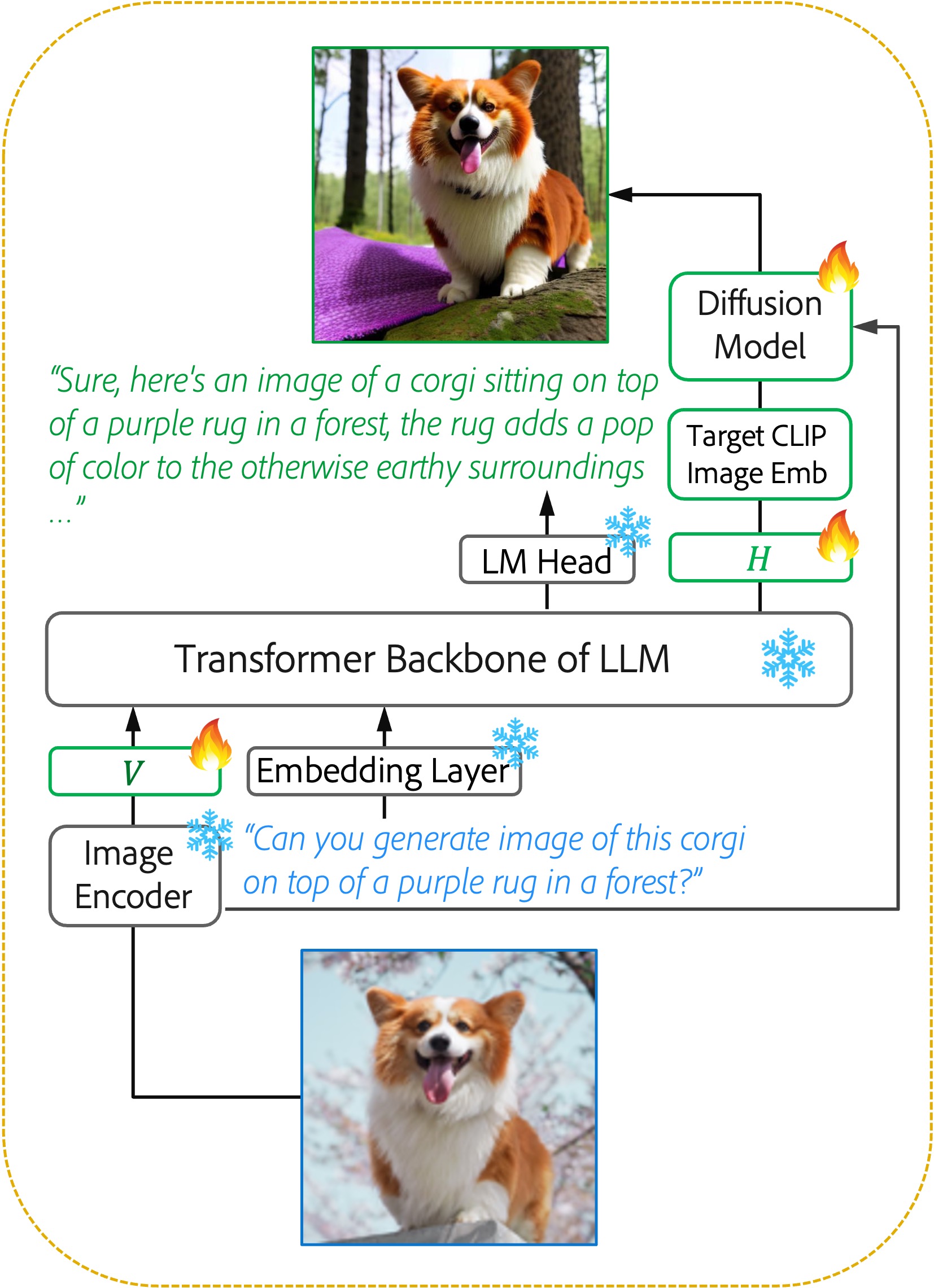}
    \vspace{-0.15in}
    \caption{Illustration of our model architecture, where the modules to be fine-tuned are indicated by flame icons. }
    \label{fig:model}
    \vspace{-0.25in}
\end{figure}

To this end, we propose to utilize the capability of large language models (LLMs). Our model architecture is presented in Figure~\ref{fig:model}, with more details discussed as follows. 

\subsection{Customization Assistant}
Let $\mathcal{E}$ be a pre-trained image encoder, $\mathcal{M}$ be our multi-modal large language model (MLLM), $\rvx$ be an image containing the object we want to generate images for. $\mathcal{E}(\rvx)$ is injected into MLLM through a vision projection layer $\text{V}$ following LLaVA~\citep{liu2023llava}.

Based on user-input image $\rvx$ and text $\rvy$, $\mathcal{M}$ will infer user's intention and outputs two sequences of embeddings: $\{\rve_i\}$ and $\{\rvs_j\}$. 
$\{\rvs_j\}$ will be mapped into special embeddings $\{\text{H}(\rvs_j)\}$ through a newly introduced projection layer $\text{H}$.  $\{\text{H}(\rvs_j)\}$ will be injected into a diffusion model $\mathcal{G}$, to guide the generation process.
Meanwhile, $\{\rve_i\}$ will be mapped into natural language through a language modelling (LM) head, which will provide additional information or explanation of the generated results.

The output embeddings $\{\text{H}(\rvs_j)\}$ are capable of capturing most of the semantic information for the target generation, while it may lose fine-grained details of the original image, due to the difficulty of aligning different modalities within single output space. Thus we also introduce $\mathcal{E}(\rvx)$ into the diffusion model $\mathcal{G}$.
Specifically,
both $\{\text{H}(\rvs_j)\}$ and $\mathcal{E}(\rvx)$ will be injected into the diffusion model through cross-attention layers:
\begin{align}\label{eq:cross_attn}
    \rvz^\prime = \text{Softmax}(\dfrac{Q_\rvz K_{\text{H}}^T}{\sqrt{d}}) V_{\text{H}} + \text{Softmax}(\dfrac{Q_\rvz K_{\mathcal{E}}^T}{\sqrt{d}}) V_{\mathcal{E}},
\end{align}
where $d$ is the scaling factor in attention mechanism, $\rvz, \rvz^\prime$ are intermediate features inside the UNet of diffusion model, $Q_\rvz$ is the query value calculated based on $\rvz$, $K_{\text{H}}, V_{\text{H}}$ denote key and value calculated based on $\{\text{H}(\rvs_j)\}$, $K_{\mathcal{E}}, V_{\mathcal{E}}$ denote key and value corresponding to $\mathcal{E}(\rvx)$.

Let $\tilde{\rvx}$ be the target generation based on input image $\rvx$ and input text $\rvy$, $\tilde{\rvy}$ denotes the target response from the language model, our MLLM $\mathcal{M}$ is trained with the loss
\begin{align}\label{eq:mllm_loss}
    \mathcal{L}_{\mathcal{M}} = \mathcal{L}_{\text{LM}}(\mathcal{M}(\rvy), \tilde{\rvy}) + \lambda d(\text{H}(\rvs), \mathcal{F}(\tilde{\rvx})),
\end{align}
where $\mathcal{L}_{\text{LM}}$ is the language modeling loss, $\mathcal{M}(\rvy)$ denotes response generated by the language model, $\mathcal{F}(\tilde{\rvx})$ denotes the CLIP image global embedding of $\tilde{\rvx}$, $d(\cdot, \cdot)$ measures the difference between two vectors, $\lambda>0$ is a hyper-parameter. In practice, we set $d(\cdot, \cdot)$ to be mean squared error, which works well than negative cosine similarity according to our experiments. We set $\lambda=0.2$ which leads to the best results in our implementation.

\begin{figure}[t!]
    \centering
    \includegraphics[width=0.9\linewidth]{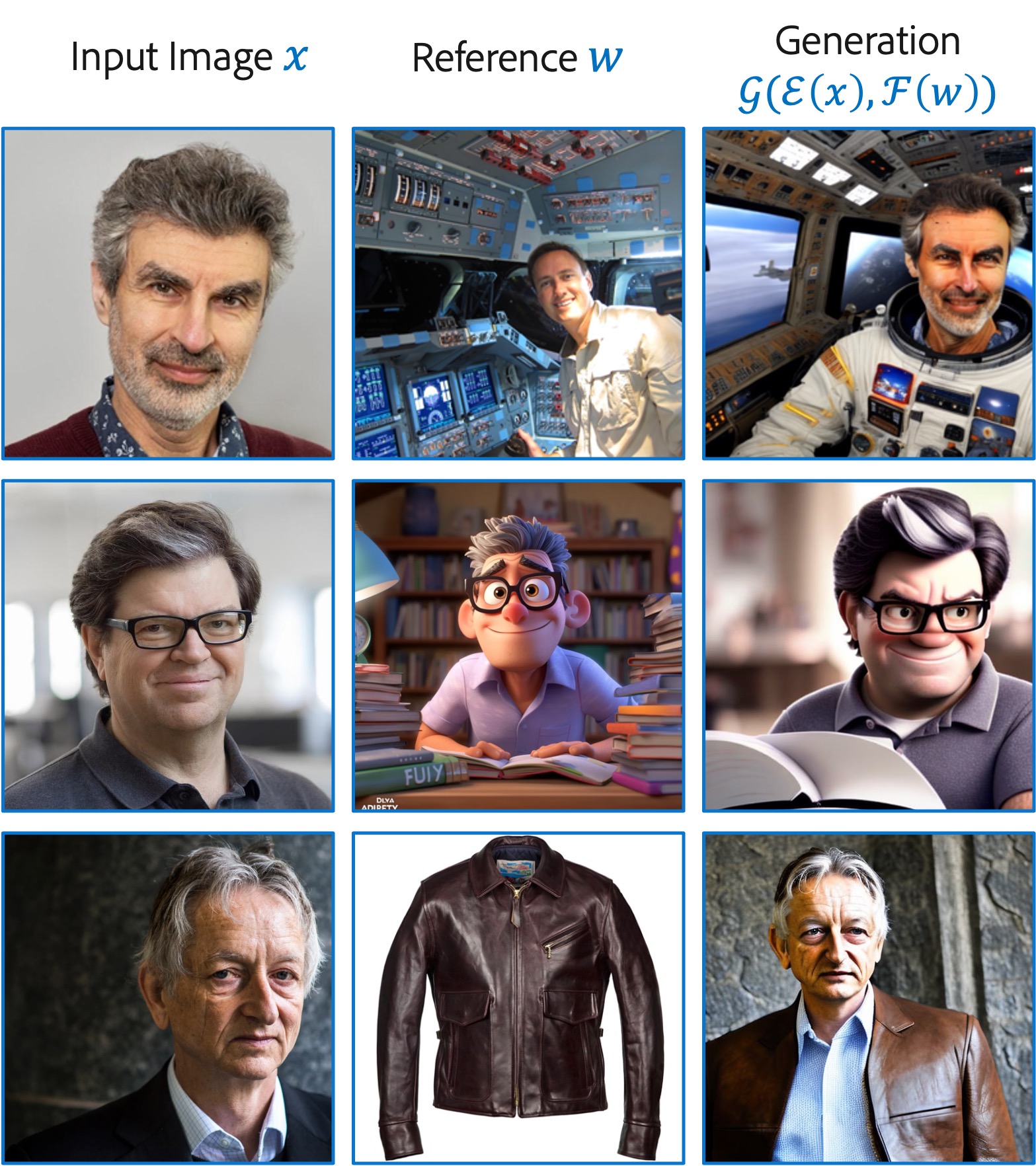}
    \caption{Our method enables tuning-free generation conditioned on multiple images. Details captured by $\mathcal{E}(\rvx)$ can be seamlessly combined with semantic $\mathcal{F}(\rvw)$. }
    \label{fig:example_condition_on_clip}
    \vspace{-0.1in}
\end{figure}

In Equation \eqref{eq:mllm_loss}, $\text{H}(\rvs)$ is designed to be a CLIP~\citep{radford2021CLIP} image embedding. This design leads to two major benefits. The first benefit is that we can train MLLM $\mathcal{M}$ and diffusion model $\mathcal{G}$ separately because diffusion model $\mathcal{G}$ does not appear in \eqref{eq:mllm_loss}. This is important especially considering the fact that both models can have billions of parameters, training MLLM and diffusion model together could be computationally prohibitive. The other important benefit is that it enables more flexible generations: when we meet difficulty in describing our target in words, we can directly use image embedding of another image $\rvw$ as semantic to guide the generation via 
\[
    \tilde{\rvx} = \mathcal{G}(\mathcal{E}(\rvx), \mathcal{F}(\rvw)),
\]
which leads to impressive results as shown in Figure~\ref{fig:example_condition_on_clip}.

Our diffusion model is trained with
\begin{align}\label{eq:loss_diffusion}
    \mathcal{L}_{\mathcal{G}} = \mathbb{E}\left[ \Vert \epsilon - \epsilon_{\theta}(\mathcal{F}(\tilde{\rvx}), \mathcal{E}(\rvx), \tilde{\rvx}_t) \Vert^2] \right]
\end{align}
where $\epsilon \sim \mathcal{N}(0, \mathbf{I})$ denotes randomly sampled noise,  $\tilde{\rvx}_t$ denotes noised sample~\citep{ho2020denoising}. 

From the above objective functions, the readers may notice that we need samples in the format of quadruplet $(\rvx, \tilde{\rvx}, \rvy, \tilde{\rvy})$ to train our model. We next present how to construct such samples for training.

\subsection{Dataset Construction}
To begin with, we prepare a collection of images $\{\rvx_i\}$ and some manually designed instructions. Then we generate target images $\{\tilde{\rvx}_{j}\}$ by randomly selecting input image and instruction and applying another customization method ProFusion~\citep{zhou2023profusion}, whose implementation is publicly available and leads to promising results efficiently. 

The generated images are then automatically filtered via CLIP and DINO similarities\footnote{We use pre-trained CLIP ViT-B/32 and DINO ViT-S/16 in all the automatic data filtering.}: we compute the image-instruction similarity and filter out image whose similarity is less than 0.3; then we filtered out image whose DINO similarity with original image is less than 0.6. After filtering, the resulting images are expected to contain target object with good identity preservation and instruction-alignment.

The resulting images will be further filtered again by human workers. The workers are provided the original image, generation instruction and generated image. They are then asked to filter out low-quality generations which are not aligned with instruction or the identity is not well preserved by human preference.

\begin{figure}[t!]
    \centering
    \includegraphics[width=0.99\linewidth]{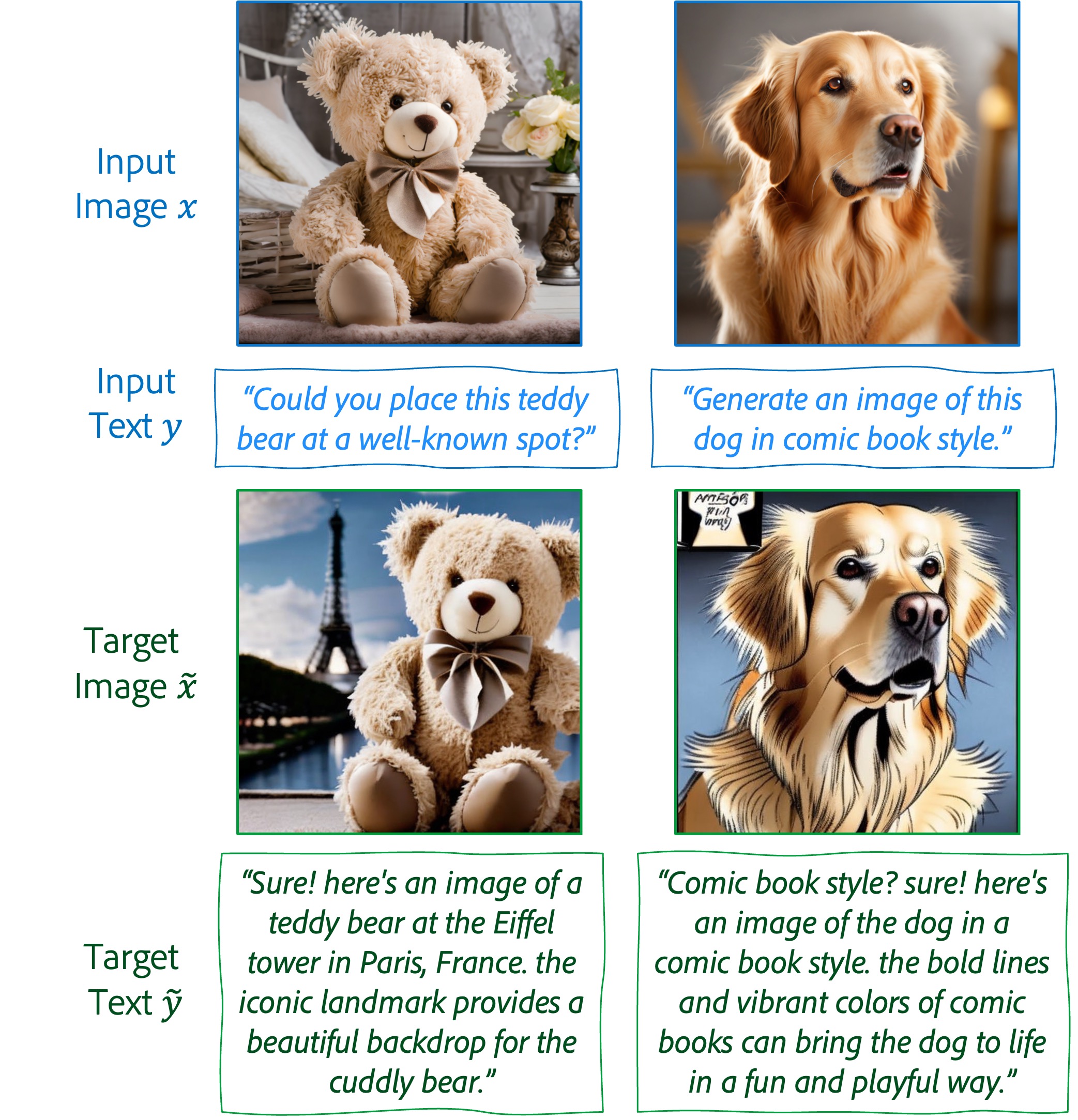}
    \caption{Two examples from our dataset, each sample contains four elements $(\rvx, \rvy, \tilde{\rvx}, \tilde{\rvy})$.}
    \label{fig:data_example}
    \vspace{-0.2in}
\end{figure}

At last, we prompt Llama-2-70B-chat~\citep{touvron2023llama2} model to generate input text $\rvy$ and target response $\tilde{\rvy}$, which simulate the interaction between user and assistant. The prompt we used is provided in the Appendix. Examples from our resulting dataset are shown in Figure~\ref{fig:data_example}.

\subsection{Self-improvement via Distillation}
Although we are able to construct desired dataset with the above pipeline, it actually requires a massive amount of computation and time cost. 

In our initial trial, 10,000 Nvidia A100 GPU hours are spent to generate around 3,000,000 samples. After automatic filtering, around 1,500 worker hours are spent to obtain the resulting dataset, which only consists of 93,000 samples.


\begin{figure*}[t!]
    \centering
    \begin{subfigure}[b]{0.99\linewidth}
        \includegraphics[width=0.99\linewidth]{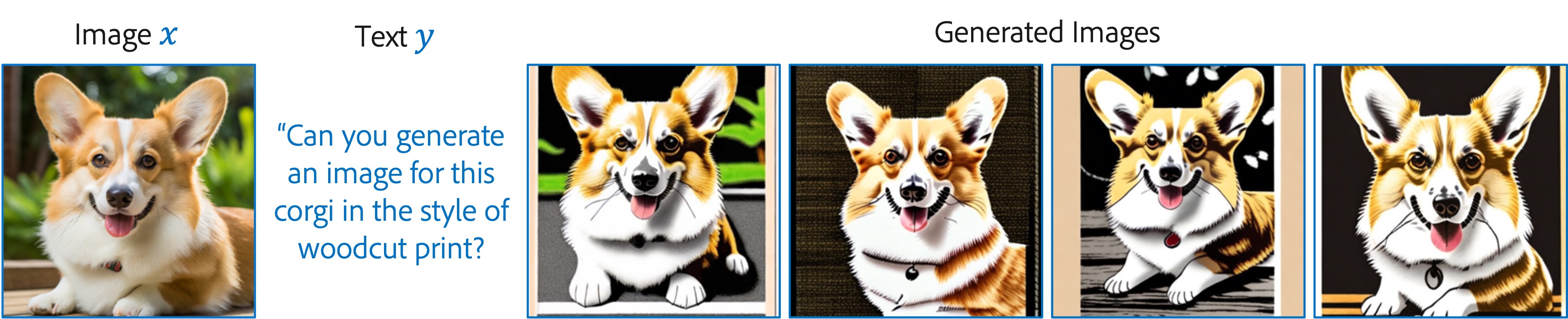}
        \caption{Generation with $\mathcal{G}(\mathcal{E}(\rvx), \text{H}(\rvs))$}
    \end{subfigure}
    \begin{subfigure}[b]{0.99\linewidth}
        \includegraphics[width=0.99\linewidth]{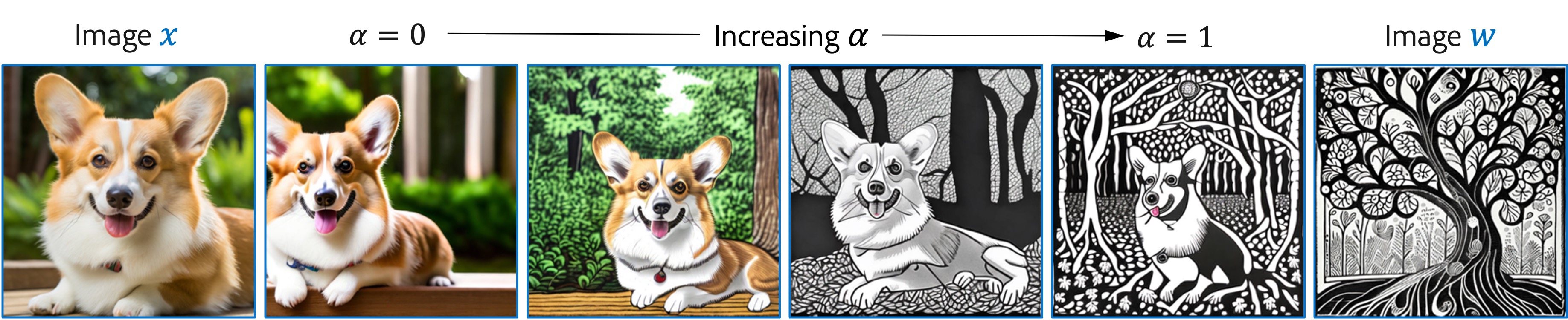}
        \caption{Generation with $\mathcal{G}(\mathcal{E}(\rvx), \alpha \mathcal{F}(\rvw) + (1 - \alpha) \mathcal{F}(\rvx))$}
    \end{subfigure}
    \vspace{-0.1in}
    \caption{We can generate high quality training data efficiently with \eqref{eq:generation}. The semantic and identity can also be easily controlled through single hyper-parameter $ \alpha$. }
    \label{fig:clip_interpolation}
    \vspace{-0.2in}
\end{figure*}


Obviously, constructing a dataset which may cover arbitrary domain is expensive with the above pipeline. 
A consequential question is, is it possible to obtain a larger and better dataset more efficiently? To this end, we propose a novel strategy, which is totally automatic, and can be easily scaled up because it does not require any human filtering workload. We present the details below.

First of all, we remove the human filtering stage in previous pipeline, and directly train a customization assistant using the automatically filtered data. Then we use the trained model to generate more samples, which will be used to fine-tune the model itself after automatic filtering, thus our strategy is termed as Self-improvement via Distillation (SID). Specifically, we propose to generate new training images by
\begin{align}\label{eq:generation}
    \tilde{\rvx} = \mathcal{G}(\mathcal{E}(\rvx), \alpha \mathcal{F}(\rvw) + (1 - \alpha) \mathcal{F}(\rvx))
\end{align}
instead of 
\[
\tilde{\rvx} = \mathcal{G}(\mathcal{E}(\rvx), \text{H}(\rvs)),
\]
where $\rvw$ is a reference image retrieved from database or generated by pre-trained Stable Diffusion with target instruction. Some details are provided in the Appendix.
As shown in Figure \ref{fig:clip_interpolation}, the initial model may fail to generate target image, when its style or semantic is rare in the training data. \eqref{eq:generation} provides an efficient way to generate those images, which can be then be used to construct a more comprehensive dataset or balance the training data distribution. Furthermore, by simply using different image $\rvw$ and hyper-parameter $\alpha$, we are able to control identity preservation and semantics efficiently. 

Recall that ProFusion requires testing time fine-tuning to perform customized generation: given a testing image, around 30 seconds of fine-tuning is needed. On the contrary, our customization assistant is a tuning-free method that generates image in few seconds, thus can save a huge amount of time. As we will show in the experiment, our proposed strategy can generate high-quality data efficiently which leads to better results even than the model trained on human filtered data.

\section{Experiment}
\label{sec:experiment}

\subsection{Implementation Details}
We conduct all the experiments with PyTorch~\citep{paszke2019pytorch} on Nvidia A100 GPUs. Our final dataset consists of around 1 million $(\rvx_i, \tilde{\rvx}_i, \rvy_i, \tilde{\rvy}_i)$ quadruplet samples, which costs around 20,000 GPU hours. 355K samples are in human image domain, while the rest images focus on open domain objects. 
The reference images $\{\rvx_i\}$ contain both public dataset and generated images: we directly use FFHQ~\citep{karras2019style} dataset to generate $(\rvx_i, \tilde{\rvx}_i, \rvy_i, \tilde{\rvy}_i)$ samples for human face domain; for object domain, we use pre-trained Stable Diffusion 2~\citep{rombach2022LDM} to generate reference images $\{\rvx_i\}$ for some selected object classes, then use the generated images to construct $(\rvx_i, \tilde{\rvx}_i, \rvy_i, \tilde{\rvy}_i)$ samples. More details are provided in the Appendix. 

Two stages are implemented in training the model: the customization assistant is first trained for 5 epochs on samples generated by ProFusion, then fine-tuned for 5 epochs on the samples generated by the first stage model. 

\begin{figure*}[ht!]
    \centering
    \includegraphics[width=0.9\linewidth]{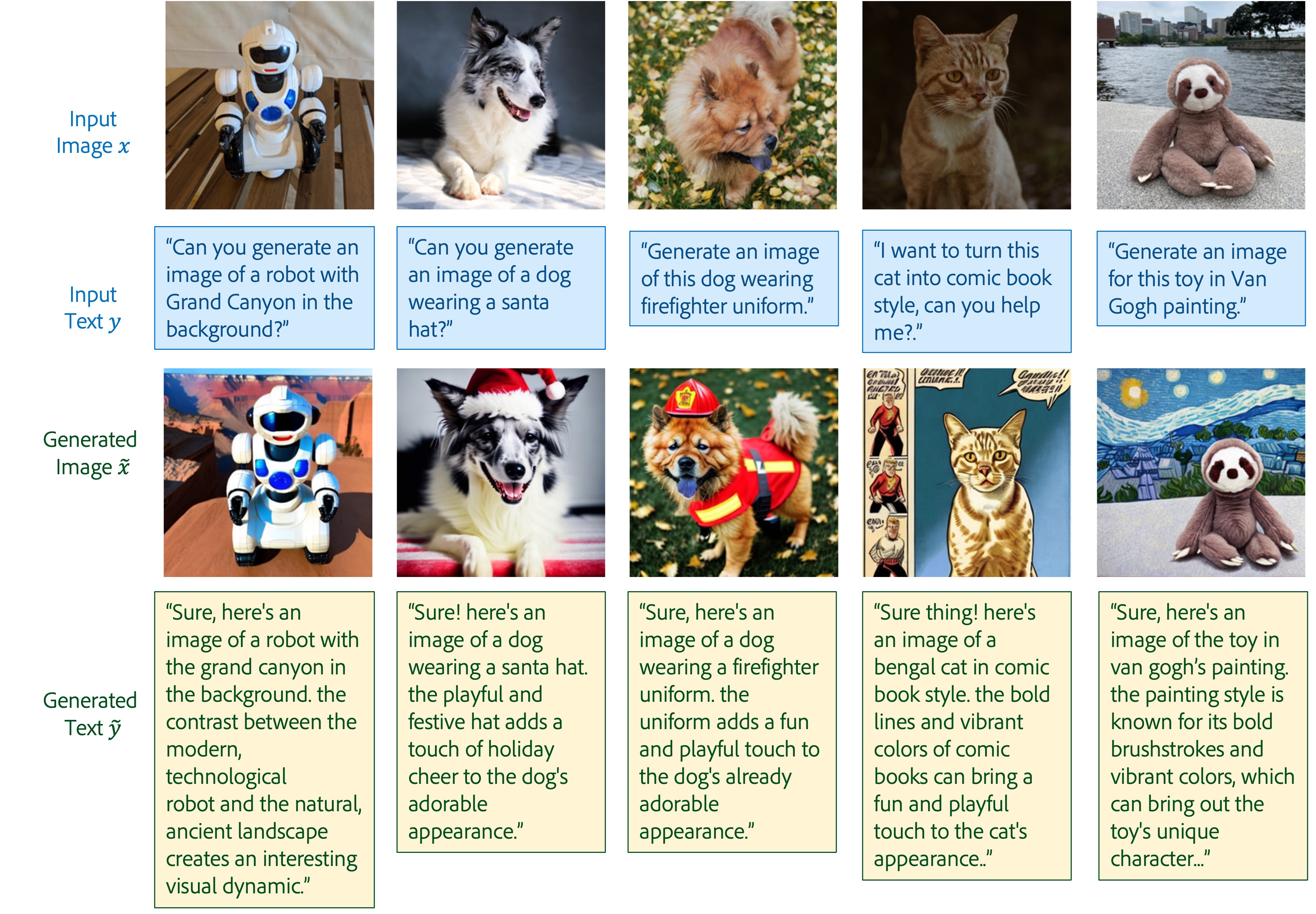}
    \caption{Generated examples from the proposed CAFE.}
    \label{fig:object_results}
    \vspace{-0.25in}
\end{figure*}

We set $\mathcal{F}$ to be CLIP ViT-L/14@336px model, and use DINOv2-Giant as image encoder $\mathcal{E}$. As a result, $\mathcal{H}(\rvs) \in \mathbb{R}^{1\times 768}, \mathcal{E}(\rvx) \in \mathbb{R}^{257\times 1536}$.
Our large language model is initialized from Llama-2-13B-chat checkpoint. 
We introduce a fully-connected layer which projects image embeddings into the input space of Llama-2, and a fully-connected layer which is the projection head for CLIP embedding prediction. A learn-able token is appended at the end of text tokens, which will be used in predicting the CLIP global embedding.
AdamW~\citep{loshchilov2018decoupled} optimizer with learning rate of 2e-3 and batch size of 128 is used. Llama-2 backbone is kept frozen, only newly introduced projection layers are fine-tuned.

Our diffusion model is initialized from pre-trained Stable Diffusion 2. 
We remove its original text encoder, and introduce new cross attention layers so that the UNet can take global embedding from CLIP ViT-L/14@336px and patch embeddings from DINOv2-Giant.
The diffusion model is also trained with AdamW optimizer. The learning rate is set to be 2e-5 and batch size is 64. During training, the DINO and CLIP embeddings are randomly dropped independently with a probability of 0.1 to enable classifier-free guidance~\citep{ho2021cfg}.


\subsection{Quantitative Results}
Following previous works~\cite{ruiz2023dreambooth,gal2023e4t}, we conduct experiments on object domain and human image domain. Some generated examples are presented in Figure~\ref{fig:object_results} and Figure~\ref{fig:main_results_human}. More examples will be provided in the Appendix, including examples on multi-round generation and image editing task.

\paragraph{Object domain} We conduct quantitative evaluation on DreamBench~\citep{ruiz2023dreambooth}, which consists of 30 subjects and 25 prompts for each subjects. 4 images are generated for each of the 750 unique combinations. Following ~\cite{ruiz2023dreambooth}, we calculate image similarities with pre-trained DINO ViT-S/16 and CLIP ViT-B/32 models, which evaluate the identity preservation between generated image and original image by computing cosine similarity between their extracted features. The metrics are denoted as DINO and CLIP-I respectively. Image-text similarity between generated image and prompt is calculated using pre-trained CLIP ViT-B/32 model, which is denoted as CLIP-T in the results.

In all the experiments, we use slightly different prompts from other methods. For example, in the case where one want to generate an image for a specific dog on the beach,
the prompt for other methods might be ``A $S^*$ dog on the beach" where $S^*$ represents the embedding capturing the characteristics of the dog. While prompt of our model is randomly selected from ``Can you generate an image for this dog on the beach?" and ``Generate an image for this dog on the beach.". Nevertheless, we still use their prompts in computing the CLIP-T similarity for fair comparison.

The main results are reported in Table~\ref{tab:dreambench}, where we compare our method with Textual Inversion~\citep{gal2022textualinversion}, DreamBooth~\citep{ruiz2023dreambooth}, CustomDiffusion~\citep{kumari2023multi}, BLIP-Diffusion~\citep{li2023blipdiffusion}, ELITE~\citep{wei2023elite}, Subject-Diffusion~\citep{ma2023subjectdiffusion}, SuTI~\citep{chen2023suti}, Kosmos-G~\citep{pan2023kosmos}. Results of baseline methods can directly taken from previous papers. Our method achieves competitive results, and is the only model which can generate text explanations and elaborations along with images.
Note that SuTI is based on Imagen~\citep{saharia2022imagen}, which is a stronger base model than our Stable Diffusion 2. SuTI requires 15-20 seconds to perform generation, while our method only needs 5 seconds, which can be further reduced to 2 seconds if one directly uses CLIP embedding from an image instead of embedding generated by MLLM. 

\paragraph{Human image domain} We then conduct experiments on human face domain following \cite{zhou2023profusion, gal2023e4t}. We train a model on human image subset of our dataset, then evaluate it with all the 23 prompts, 7 researcher images provided in \cite{gal2023e4t}. 10 images are generated for each image-prompt combination. The generated images are then evaluated by two metrics following ~\cite{zhou2023profusion}: we utilize CLIP ViT-B/32 models to calculate the image-text similarity; we evaluate identity similarity by the cosine similarity between extracted features of generated and original image using pre-trained face recognition model~\citep{kim2022adaface}.  The main results are presented in Table~\ref{tab:face_clip} where the identity similarity is denoted as ID. We compare our method with previous methods including Textual Inversion~\citep{gal2022textualinversion}, DreamBooth~\citep{ruiz2023dreambooth}, E4T~\citep{gal2023e4t} and ProFusion~\citep{zhou2023profusion}. Better results are obtained with our method, indicating the effectiveness of the proposed method. Some qualitative comparisons are provided in Figure~\ref{fig:main_results_human}.  We also include results from another tuning-free method PhotoVerse~\citep{chen2023photoverse}, which is specifically designed for human face domain. However, the implementation of PhotoVerse is not available, thus only qualitative comparison is provided.  We can see that the proposed CAFE leads to better identity preservation and image fidelity.

\begin{figure*}[t!]
    \centering
    \includegraphics[width=0.99\linewidth]{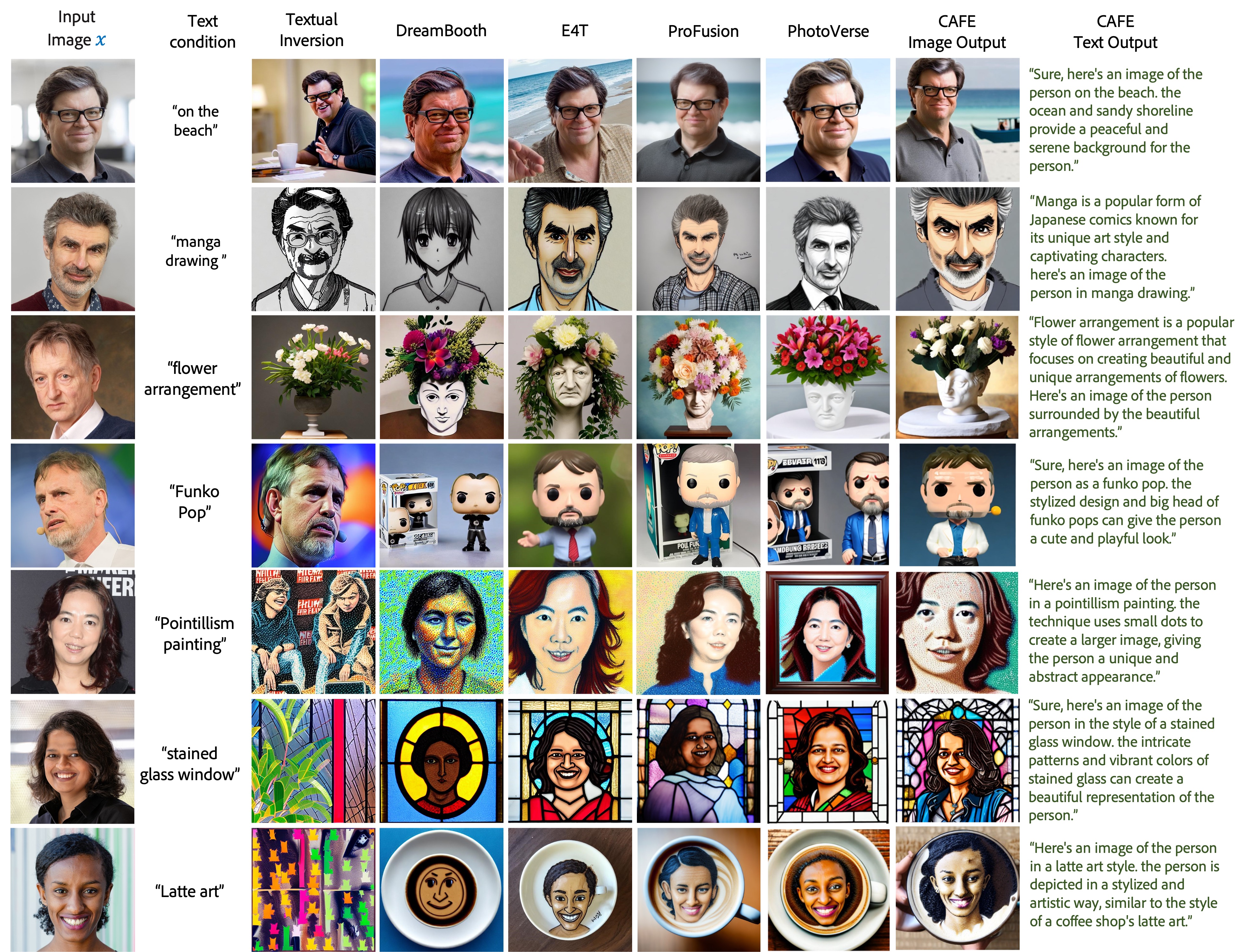}
    \caption{Comparison with related methods on human images. Results of other methods are directly taken from corresponding papers. PhotoVerse and CAFE are tuning-free methods. Our CAFE is able to capture details such as the tiny microphone in the Funko Pop example.}
    \label{fig:main_results_human}
    \vspace{-0.2in}
\end{figure*}
\begin{table}[t!]
    \centering
    \scalebox{0.8}{
    \begin{tabular}{lcccc}
    \toprule
        Method & Tuning-free  & DINO $(\uparrow)$ & CLIP-I $(\uparrow)$ & CLIP-T $(\uparrow)$ \\
        \midrule
        Real Images & - & 0.774 & 0.885 & - \\
        \midrule
        Textual Inversion & \xmark & 0.569 & 0.780 & 0.255 \\
        DreamBooth & \xmark & 0.668 & 0.803 & 0.305 \\
        CustomDiffusion & \xmark & 0.643 & 0.790 & 0.305 \\
        BLIP-Diffusion & \xmark & 0.670 & 0.805 & 0.302 \\
        \midrule
        BLIP-Diffusion & \cmark & 0.594 & 0.779 & 0.300 \\
        Re-Imagen & \cmark & 0.600 & 0.740 & 0.270 \\
        ELITE & \cmark & 0.621 & 0.771 & 0.293 \\
        Subject-Diffusion & \cmark & 0.711 & 0.787 & 0.293 \\
        SuTI &\cmark & 0.741 & 0.819 & 0.304 \\
        Kosmos-G & \cmark & 0.694 & 0.847 & 0.287 \\
        CAFE (Ours) & \cmark & 0.715 & 0.827 & 0.294\\
        \bottomrule
    \end{tabular}
    }
    \vspace{-0.1in}
    \caption{Quantitative evaluation on DreamBench.}
    \label{tab:dreambench}
    \vspace{-0.1in}
\end{table}

\begin{table}[t!]
    \centering
    \scalebox{0.8}{
    \begin{tabular}{lccc}
        \toprule
        Method & Tuning-free & ID $(\uparrow)$ & CLIP-T $(\uparrow)$\\
        \midrule
        Textual Inversion & \xmark& 0.210 & 0.257 \\
        Dreambooth & \xmark & 0.307 & 0.283\\
        E4T & \xmark & 0.426 & 0.277 \\
        ProFusion &\xmark & 0.432 & 0.293\\
        CAFE (Ours) & \cmark & 0.464 & 0.297\\
        \bottomrule
    \end{tabular}
    }
    \vspace{-0.1in}
    \caption{Results evaluated in human face domain.}
    \label{tab:face_clip}
    \vspace{-0.2in}
\end{table}

\subsection{Ablation Studies}
\paragraph{Effectiveness of SID}
One important question is, given the same amount of computation resources and time, will the proposed SID training strategy lead to a better model which outperforms the model trained on human filtered data samples?

We conduct an ablation study to answer the above question. Specifically, we start from a dataset generated by ProFusion in human face domain, which cost around 10,000 A100 GPU hours. Then different model variants are trained on the following datasets:

\begin{itemize}
    \item Dataset $\mathcal{D}_1$, which only contains automatically filtered samples;
    \item Dataset $\mathcal{D}_2$, which is constructed by asking human workers to filter out low-quality samples in $\mathcal{D}_1$;
    \item Dataset $\mathcal{D}_3$, which only consists of samples generated by the model trained on $\mathcal{D}_1$. Filtering with CLIP and DINO similarity is also performed;
    \item Dataset $\mathcal{D}_4$, which is the union $\mathcal{D}_1 \cup \mathcal{D}_3$;
\end{itemize}

The cost of generating samples for $\mathcal{D}_3$ is set to be 1,500 GPU hours for fair comparison, as 1,500 worker hours are spent in human filtering stage of constructing $\mathcal{D}_2$. 

The comparison is provided in Table~\ref{tab:different_dataset}, along with some dataset statistics. All the models use the same architecture. From the results we can conclude the proposed self-distillation strategy does lead to better performance. Although human filtered dataset may have better quality in terms of image fidelity, it may not lead to a model with good generalization ability because the amount of sample is limited. We also notice that the model trained on $\mathcal{D}_3$ leads to good performance, illustrating the effectiveness of constructing dataset with \eqref{eq:generation}.

\begin{table}[t!]
    \centering
    \scalebox{0.8}{
    \begin{tabular}{ccccc}
        \toprule
        Dataset & Total Cost & $\#$ of Sample  & ID $(\uparrow)$ & CLIP-T $(\uparrow)$\\
        \midrule
        $\mathcal{D}_1$ & 10,000 GPU hour & 207K & 0.433 & 0.294\\
        $\mathcal{D}_2$ & 11,500 hour & 93K& 0.441 & 0.288\\
        $\mathcal{D}_3$ & 11,500 GPU hour & 148K & 0.465 & 0.291\\
        $\mathcal{D}_4$ & 11,500 GPU hour & 355K& 0.464 & 0.297\\
        \bottomrule
    \end{tabular}
    }
    \vspace{-0.1in}
    \caption{Ablation study with different dataset, the proposed can efficiently construct a high-quality dataset, which leads to improved model performance.}
    \label{tab:different_dataset}
    \vspace{-0.1in}
\end{table}

\paragraph{Different objective functions}
As mentioned in Section~\ref{sec:method}, we can choose different $d(\cdot, \cdot)$ in \eqref{eq:mllm_loss}. Because the CLIP model is trained with contrastive loss using cosine similarity, thus we conduct ablation study to compare using negative cosine similarity and mean squared error in \eqref{eq:mllm_loss}. Hyper-parameter $\lambda$ in \eqref{eq:mllm_loss} is selected from $\left[0.1, 0.2, 0.5, 1.0, 2.0\right]$ by their resulting performance. 
The quantitative evaluation is presented in Table \ref{tab:ablation_mse_cos}, from which we can see that mean square error leads to better performance. 

\begin{table}[t!]
    \centering
    \scalebox{0.9}{
    \begin{tabular}{lcc}
        \toprule
        Loss Function  & ID $(\uparrow)$ & CLIP-T $(\uparrow)$\\
        \midrule
        Mean Sqaure Error & 0.464 & 0.297\\
        Nagative Cosine Similarity & 0.456 & 0.295 \\
        \bottomrule
    \end{tabular}
    }
    \vspace{-0.1in}
    \caption{Ablation study with different loss functions.}
    \label{tab:ablation_mse_cos}
\end{table}

\begin{table}[t!]
    \centering
    \scalebox{0.8}{
    \begin{tabular}{llcc}
        \toprule
        CLIP Model & DINO Model & ID $(\uparrow)$ & CLIP-T $(\uparrow)$\\
        \midrule
         ViT-B/32 & DINOv2-Giant &  0.449 & 0.252\\
         ViT-L/14@336px & DINOv2-Base & 0.408 & 0.295 \\
         \midrule
         ViT-L/14@336px & DINOv2-Giant & 0.464 & 0.297 \\
        \bottomrule
    \end{tabular}
    }
    \vspace{-0.1in}
    \caption{Ablation study with different image encoders.}
    \label{tab:ablation_different_encoder}
    \vspace{-0.2in}
\end{table}

\paragraph{Different image encoders}
Recall that CLIP ViT-L/14@336px and DINOv2-Giant are used in our experiments, readers may be curious about how will different variants of these pre-trained models influence the model performance. 
To better understand the impact of different image encoders, we conduct ablation study where the encoders are replaced by smaller variants. The quantitative results are presented in Table~\ref{tab:ablation_different_encoder}, which is also evaluated on human faces following previous experiments. 
As expected, model with CLIP ViT-B/32 encoder obtains much worse results in terms of image-text similarity; while DINOv2-base leads to worse identity similarity than DINOv2-Giant.

\paragraph{Contribution of CLIP and DINO embedding}
We also conduct ablation study where only CLIP embeddings or DINO embeddings are used in generation. The results are presented in Table~\ref{tab:ablation_different_path}, where we can find that using only CLIP embedding leads to good CLIP-T score and bad ID score, while using only DINO embedding leads to good ID score and poor CLIP-T score. The results are aligned with our expectation that semantics are mainly controlled by CLIP embedding, fine-grained details are mainly controlled by DINO embedding. 

\begin{table}[t!]
    \centering
    \scalebox{0.8}{
    \begin{tabular}{llcc}
        \toprule
        CLIP embedding & DINO embedding & ID $(\uparrow)$ & CLIP-T $(\uparrow)$\\
        \midrule
         \cmark & \xmark &  0.216 & 0.297 \\
         \xmark & \cmark &  0.418 & 0.234\\
         \midrule
         \cmark & \cmark & 0.464 & 0.297 \\
        \bottomrule
    \end{tabular}
    }
    \vspace{-0.1in}
    \caption{Ablation study where only CLIP embedding or DINO embedding is used in generation.}
    \label{tab:ablation_different_path}
    \vspace{-0.2in}
\end{table}
\section{Conclusion}
\label{sec:conclusion}
In this work, we propose CAFE which is a tuning-free method for customizing pre-trained text-to-image generation model. Different from existing works, the proposed CAFE is based on large language model thus can handle ambiguous user input and output explanations along with generated images. A novel strategy is proposed, which leads to more efficient and scalable dataset construction for training better CAFE. Competitive results are obtained in experiments across different domains, indicating the effectiveness of the proposed method. 
\clearpage
{
    \small
    \bibliographystyle{ieeenat_fullname}
    \bibliography{main}
}

\clearpage
\appendix
\onecolumn

\section{Implementation Details}

We provide some implementation details here.

\begin{figure*}[h!]
    \centering
    \includegraphics[width=0.99\linewidth]{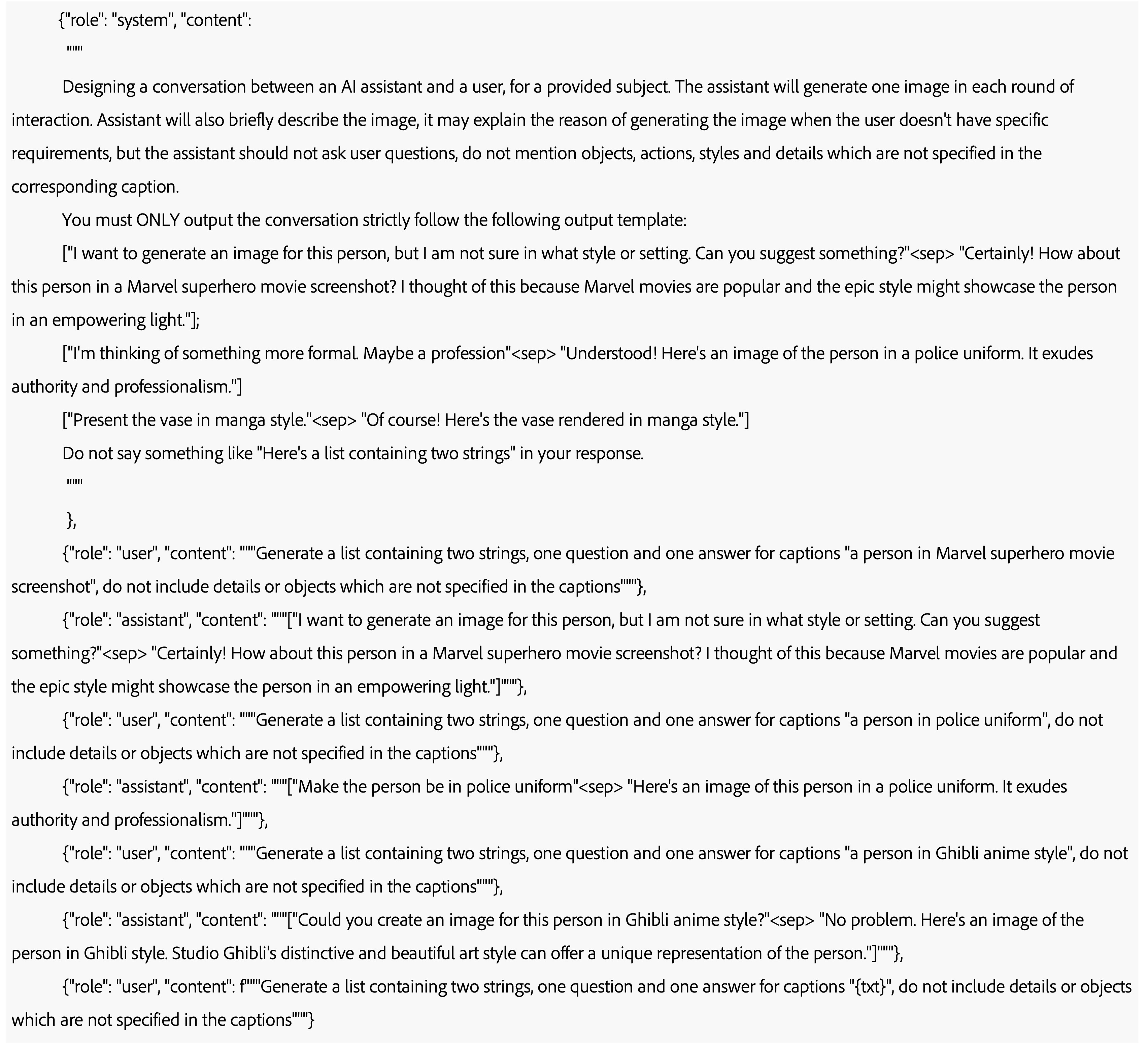}
    \caption{Prompt example we used to generate data samples with Llama-2-70B-chat model.}
    \label{fig:llama_prompt}
\end{figure*}

\paragraph{Text data generation} We provide prompt example which we use to generate text data with the Llama-2-70B-chat model in Figure~\ref{fig:llama_prompt}. We are able to obtain the expected output with this prompt in most cases, while it also leads to failure sometimes. Thus we apply an automatic filtering mechanism on the generated text data, where undesired outputs such as empty strings are filtered out.

\begin{figure*}[ht!]
    \centering
    \includegraphics[width=0.9\linewidth]{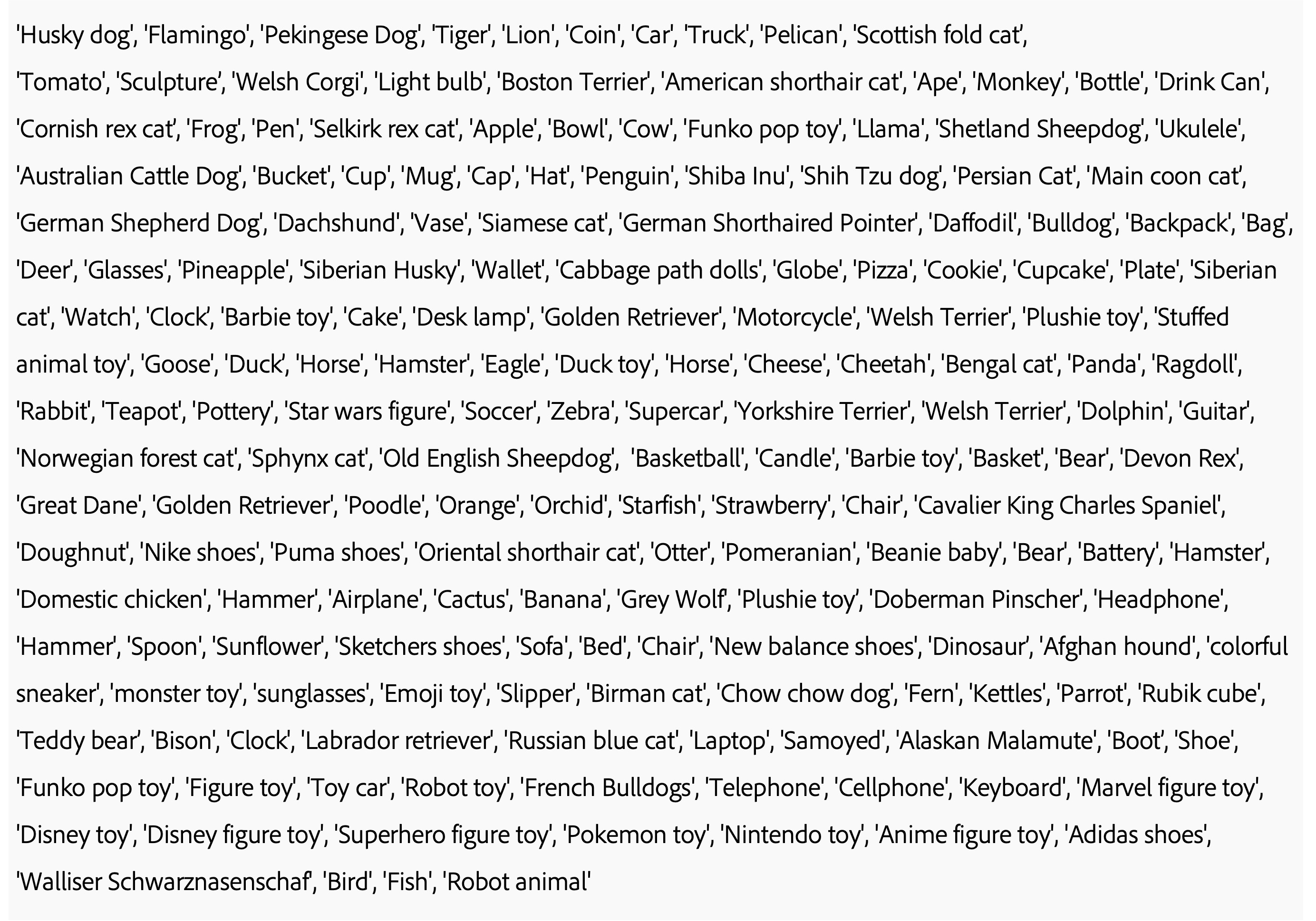}
    \caption{Classes we used to generate original images for object domain.}
    \label{fig:classes}
\end{figure*}

\paragraph{Image data generation} As mentioned in Section~\ref{sec:experiment}, original images $\{\rvx_i\}$ in object domain are generated first generated with pre-trained text-to-image generation model (Stable Diffusion). Let $c_i$ be a class name, we use prompt ``A realistic photo picture of $c_i$, ultra quality, sharp focus, tack sharp, dof, film grain, Fujifilm XT3, crystal clear, 8K UHD, highly detailed" to generate the image. $c_i$ is randomly selected from the classes shown in Figure~\ref{fig:classes}.

\begin{figure*}[ht!]
    \centering
    \includegraphics[width=0.9\linewidth]{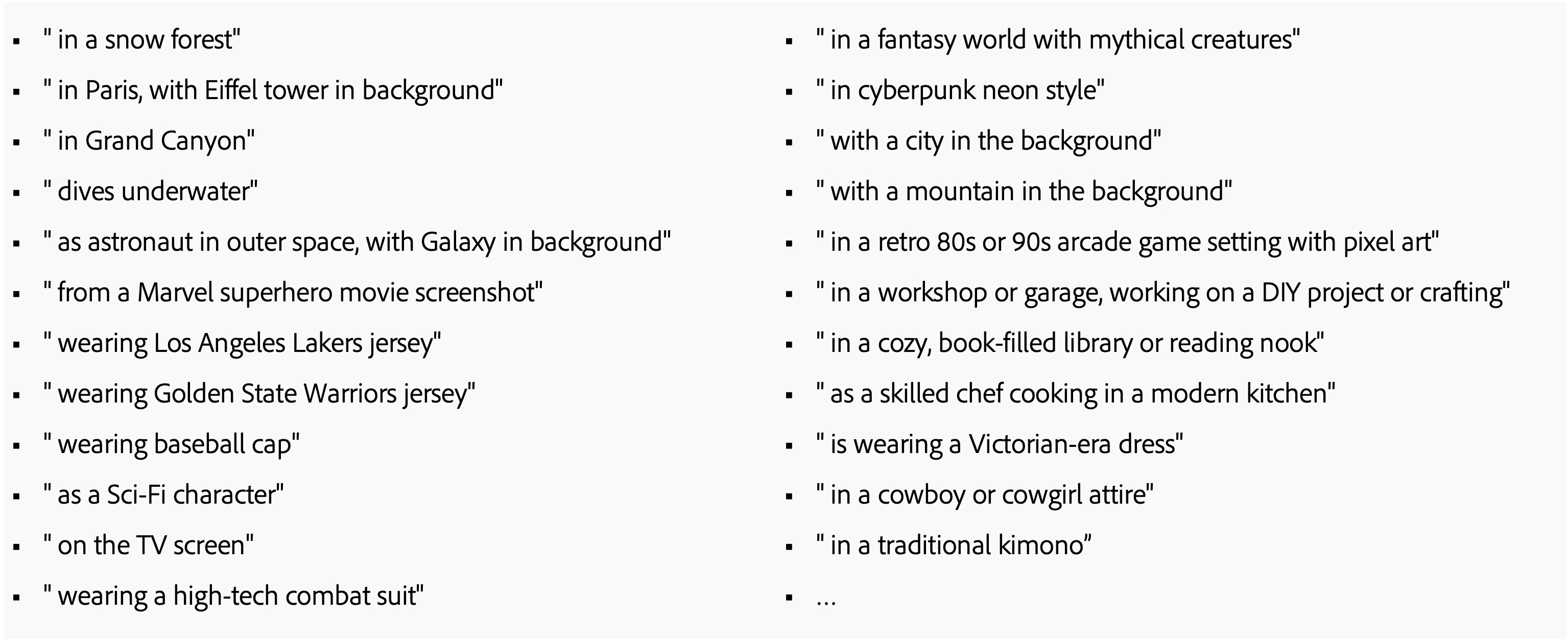}
    \caption{Examples of instructions we used in generating target images.}
    \label{fig:instruction}
\end{figure*}

With the original images, manually designed instructions are used to generate the target images $\{\tilde{\rvx}_i\}$ with ProFusion. Examples of our designed instructions are shown in Figure~\ref{fig:instruction}

When we generate data samples with the assistant trained on automatically filtered data. We first generate images $\rvw$ with pre-trained Stable Diffusion, then apply Equation \eqref{eq:generation} with $\alpha=0.7$.

\section{Generated Examples}
Some generated examples are provided below. We also provide examples where the proposed method is applied in multi-round generation and image editing task.

\begin{figure*}[ht!]
    \centering
    \includegraphics[width=0.8\linewidth]{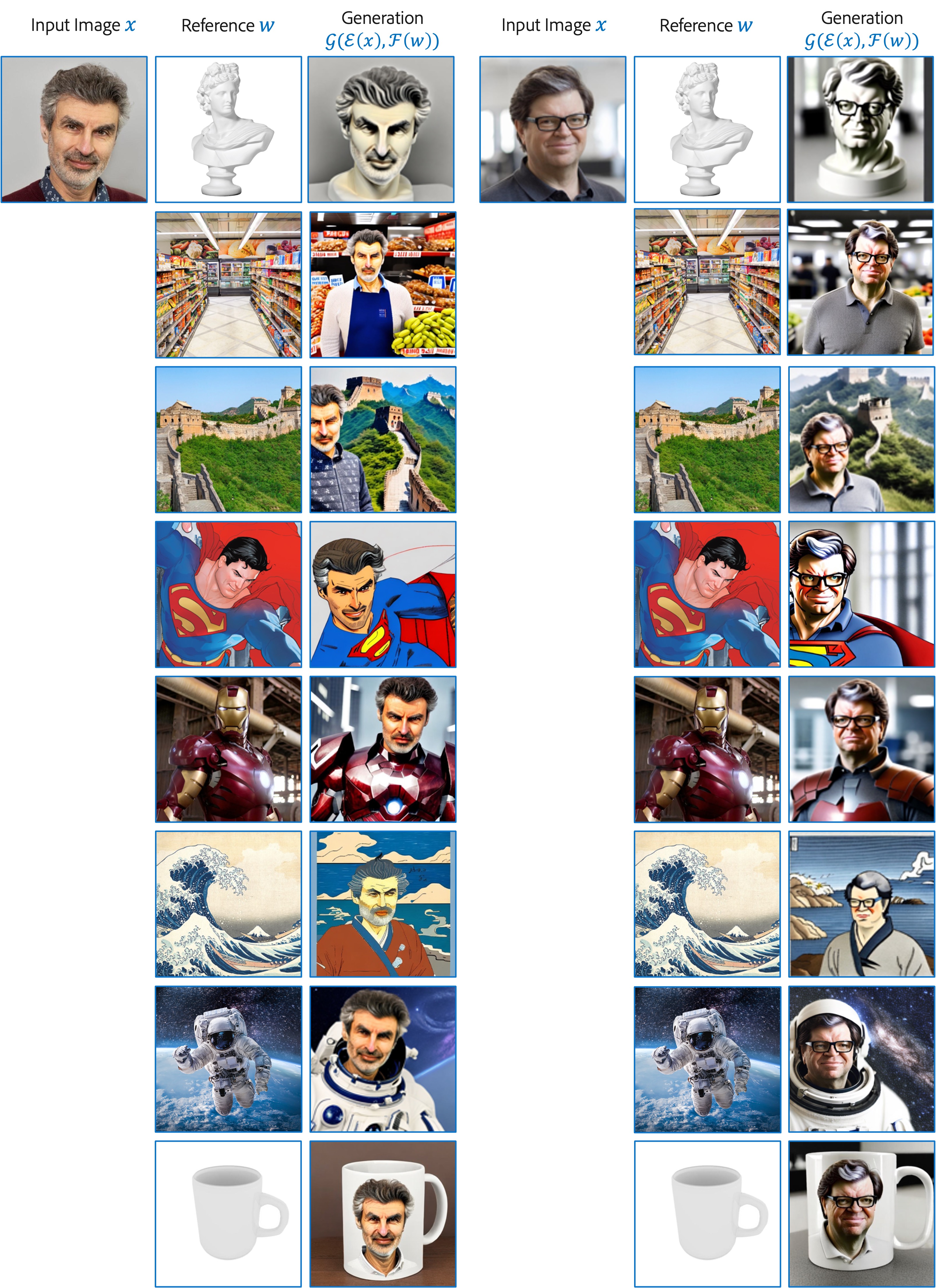}
    \caption{Generated examples from the proposed CAFE.}
    \label{fig:gen_examples}
\end{figure*}

\begin{figure*}[t!]
    \centering
    \includegraphics[width=0.5\linewidth]{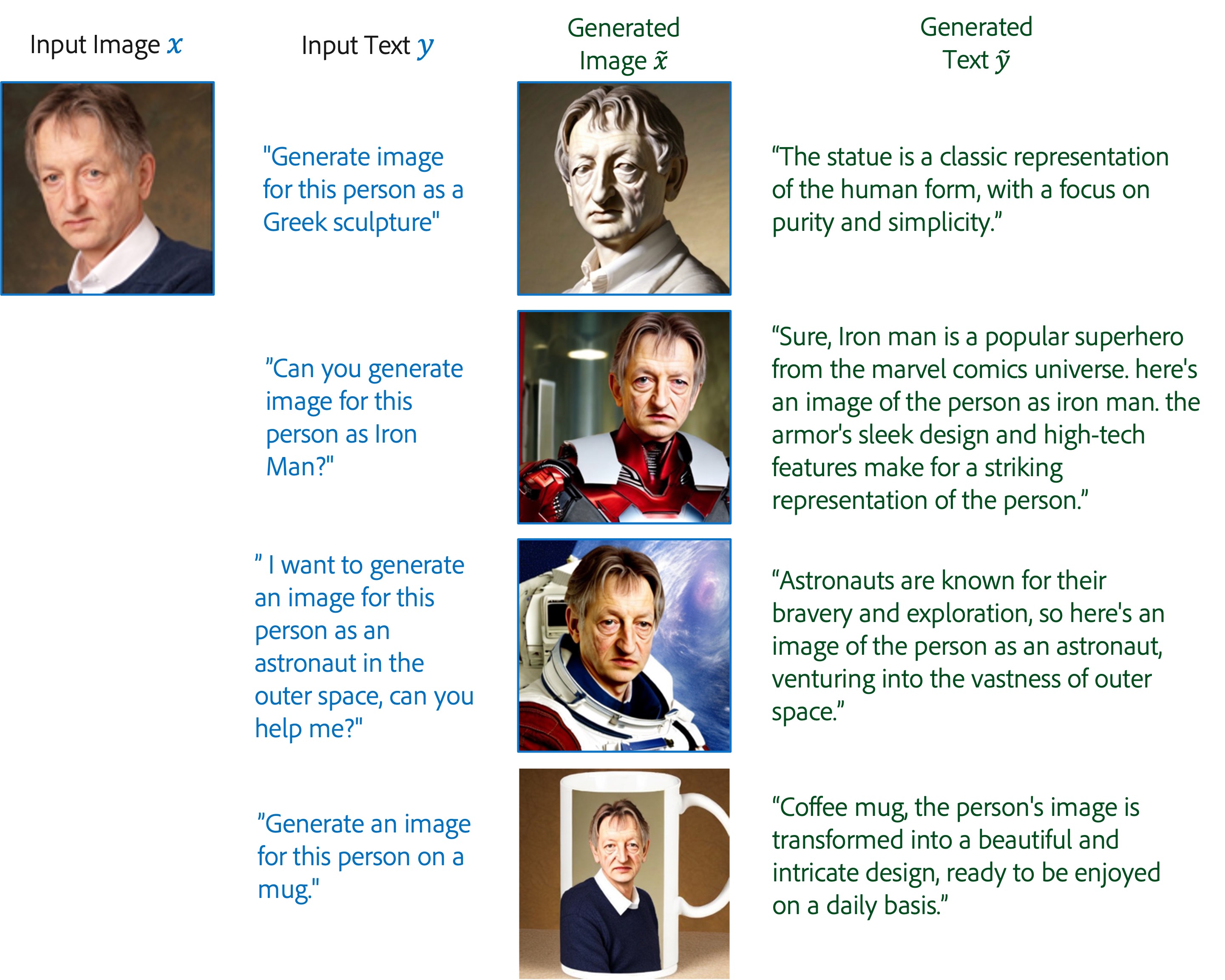}
    \caption{Generated examples from the proposed CAFE.}
    \label{fig:gen_examples_2}
\end{figure*}

\begin{figure*}[t!]
    \centering
    \includegraphics[width=0.5\linewidth]{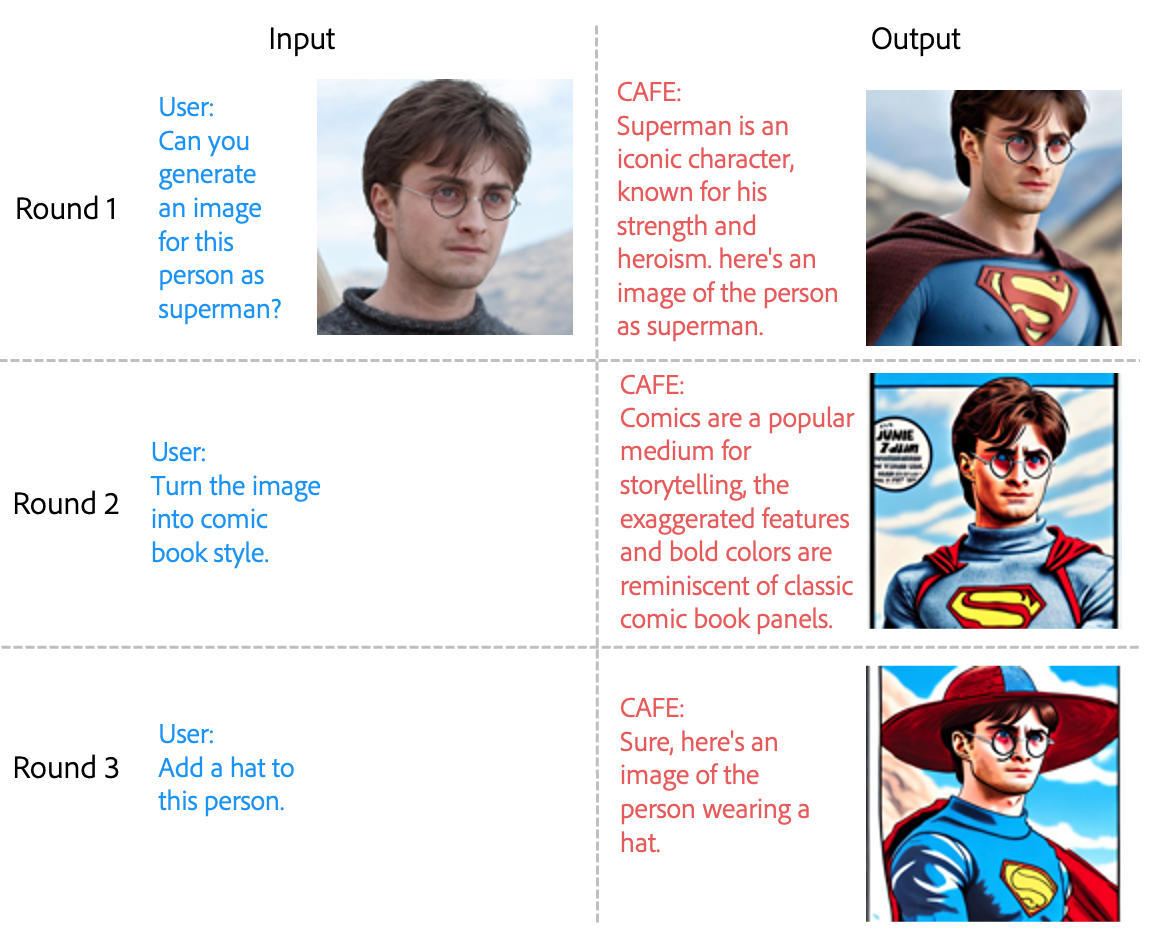}
    \caption{Multi-round generation examples from the proposed CAFE.}
    \label{fig:multi-round}
\end{figure*}

\begin{figure*}[t!]
    \centering
    \includegraphics[width=0.5\linewidth]{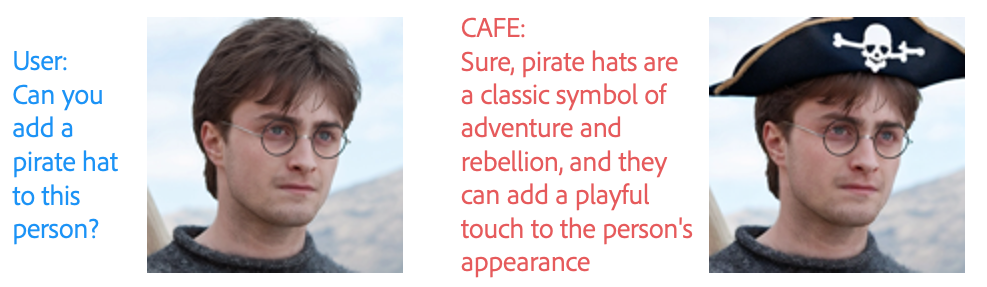}
    \caption{Image editing examples from the proposed CAFE.}
    \label{fig:editing}
\end{figure*}


\end{document}